\documentclass[sigconf,nonacm]{acmart}

\acmConference[EVIA '23]{Make sure to enter the correct
  conference title from your rights confirmation emai}{December 12--15,
  2023}{Tokyo, JP}

\acmBooktitle{The Tenth International Workshop on Evaluating Information Access (EVIA 2023),
  December 12--15, 2023, Tokyo, JP}

\acmSubmissionID{57}

\begin{document}

\title{Heaps' Law in GPT-Neo Large Language Model Emulated Corpora}

\author{Uyen Lai}
\orcid{0009-0007-9802-915X}
\email{tlai11167@upei.ca}
\affiliation{%
  \institution{University of Prince Edward Island}
  \streetaddress{550 University Ave}
  \city{Charlottetown}
  \state{Prince Edward Island}
  \country{Canada}
  \postcode{C1A 4P3}
}

\author{Gurjit S. Randhawa}
\orcid{0000-0003-1054-125X}
\email{grandhawa@upei.ca}
\affiliation{%
  \institution{University of Prince Edward Island}
  \streetaddress{550 University Ave}
  \city{Charlottetown}
  \state{Prince Edward Island}
  \country{Canada}
  \postcode{C1A 4P3}
}

\author{Paul Sheridan}
\orcid{0000-0002-5484-1951}
\authornote{Corresponding author}
\email{paul.sheridan.stats@gmail.com}
\affiliation{%
  \institution{University of Prince Edward Island}
  \streetaddress{550 University Ave}
  \city{Charlottetown}
  \state{Prince Edward Island}
  \country{Canada}
  \postcode{C1A 4P3}
}

\renewcommand{\shortauthors}{Lai, Randhawa, and Sheridan}

\begin{abstract}
Heaps' law is an empirical relation in text analysis that predicts vocabulary growth as a function of corpus size. While this law has been validated in diverse human-authored text corpora, its applicability to large language model generated text remains unexplored. This study addresses this gap, focusing on the emulation of corpora using the suite of GPT-Neo large language models. To conduct our investigation, we emulated corpora of PubMed abstracts using three different parameter sizes of the GPT-Neo model. Our emulation strategy involved using the initial five words of each PubMed abstract as a prompt and instructing the model to expand the content up to the original abstract's length. Our findings indicate that the generated corpora adhere to Heaps' law. Interestingly, as the GPT-Neo model size grows, its generated vocabulary increasingly adheres to Heaps' law as as observed in human-authored text. To further improve the richness and authenticity of GPT-Neo outputs, future iterations could emphasize enhancing model size or refining the model architecture to curtail vocabulary repetition.
\end{abstract}


\begin{teaserfigure}
  \includegraphics[width=\textwidth]{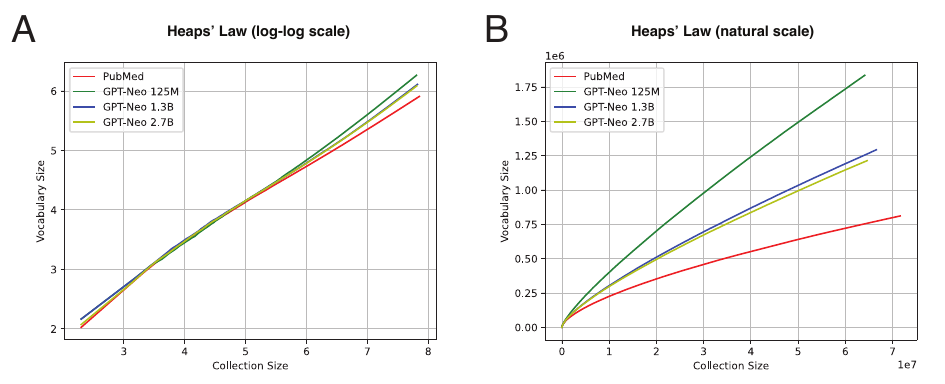}
  \caption{Heaps' law for the PubMed Abstracts corpus and corpora of GPT-Neo emulated PubMed abstracts on (A) a log-log scale in base~10, and~(B) a natural scale.}
  \label{fig:heaps-law}
\end{teaserfigure}


\maketitle

\section{Introduction}
Heaps' Law~\cite{Heaps1978, Herdan1960}, named after Harold Stanley Heaps, is a well-established empirical law in the field of quantitative linguistics. The law describes the relationship between vocabulary size (a.k.a, number of distinct words) and collection size (a.k.a, total number of terms counting multiplicities) in text. For a corpus of $d\geq 1$ documents, this relationship can be mathematically represented as
\begin{equation*} \label{eq:heaps-law}
V(N_i) = \alpha N_i^\beta
\end{equation*}
where $\alpha>0$ and $\beta>0$ are parameters to be estimated. In this paper, the following notation will be employed to denote specific characteristics of a corpus of interest. Let \( n_i \) represent the total number of terms in the \( i \)'th document, with the range of \( i \) spanning from 1 to~\( d \), where \( d \) is the total number of documents under consideration. The terms here are counted with multiplicities. We further define~\( N_i \) as the cumulative count of terms across the first~\( i \) documents, expressed as \( N_i = \sum_{\ell=1}^i n_\ell \). Additionally, \( V(N_i) \) is introduced to represent the vocabulary size of the first \( i \) documents within the corpus.

Heaps' law has been corroborated across a multitude of real-world corpora~\cite{Chacoma2020}. However, there remains a lacuna in its application and verification pertaining to text generated by large language models (LLMs). This manuscript endeavors to bridge this knowledge gap by delving into this very investigation.

In our investigation, we employed a spectrum of GPT-Neo models~\cite{Black2021}, from the most compact to the most expansive in terms of parameter count, to produce textual outputs using initial seeds of five terms from human-crafted PubMed abstracts. The overarching goal was to analyze the richness of vocabulary in LLM-generated text relative to its human counterpart. Leveraging Heaps' law for analyzing vocabulary expansion, we discerned a compelling pattern: the growth in GPT-Neo model size brings its vocabulary generation progressively in line with human linguistic behavior, characterized by a reduction in distinct word usage. This observation implies that for improved linguistic variety and closer alignment to human expression in GPT-Neo outputs, forthcoming advancements might emphasize increasing model size or refining its structure to minimize vocabulary repetitiveness.

GPT-Neo generated corpora are in keeping with Heaps’ law but exhibit a heightened vocabulary growth, as compared with PubMed abstract from which they were emulated, largely attributable to the generation of singleton terms. Furthermore, as GPT-Neo models advance in their sophistication, they begin to more adeptly mirror human linguistic tendencies.

In summary: GPT-Neo's outputs conform to Heaps’ law but demonstrate an amplified vocabulary growth, primarily driven by singleton term generation. Moreover, advanced GPT-Neo models more closely emulate human linguistic patterns. In essence, while LLM text aligns with Heaps’ law, simpler models showcase a vocabulary growth surpassing typical human-authored texts.

The implications of our findings have potential applications in the realm of LLMs. One of the pivotal applications lies in the judicious setting of parameters for these models. Instead of arbitrarily inflating the model size, our results indicate the value of optimizing the number of parameters, thus ensuring efficient performance without overkill. Such a balanced approach not only enhances computational efficiency but also offers a sustainable pathway for future LLM development.

\section{Related Work}
Empirical studies have consistently reinforced the application of Heaps' law across diverse textual domains: from novels~\cite{Font2013}, expansive literary collections encompassing 75 distinct texts~\cite{Chacoma2020}, to large-scale repositories like the Google Ngram corpus~\cite{Gerlach2012, Gerlach2014}, English Wikipedia~\cite{Gerlach2014}, PLoS ONE journals~\cite{Gerlach2014}, and the Gutenberg corpus~\cite{Tria2018}. Chacoma~et~al.~\cite{Chacoma2020} emphasized the frequent investigation of Heaps' law in concatenated compilations such as Project Gutenberg and Google Books, spotlighting its versatility~\cite{Chacoma2020}. In more expansive datasets, Williams and Zobel~\cite{Williams2005} reiterated the consistent growth patterns observed in earlier research endeavours. Delving into the intricacies of text generation, Baeza-Yates and Navarro~\cite{BaezaYates2004} showcased the adherence of certain finite state machines, reminiscent of Miller’s monkey, to both Zipf’s and Heaps’ laws, illuminating a profound interconnectedness between their respective exponents~\cite{BaezaYates2004, Hawking2020}. Hawking~et~al.~\cite{Hawking2020} explored the potential of simulated corpora in the realm of information retrieval, offering a structured methodology for experimental evaluation~\cite{Hawking2020}. Furthermore, Hawking~et~al.~\cite{Hawking2020} articulated the inherent constraints of word-based Markov generators like MarkovGenerator. Such mechanisms are bound by the vocabulary breadth of their training corpus, leading to potential Heaps' law aberrations in scaled scenarios, thereby highlighting the utility of utilizing a comprehensive subset of training data~\cite{Hawking2020}.

\section{Experiments and Results}
To investigate the validity of Heaps' law in LLM generated corpora, we emulated PubMed abstracts using the GPT-Neo suite of models. In particular, our investigations are motivated by the research questions:
\begin{itemize}
    \item Is Heaps' law valid for LLM generated text?
    \item If valid, how, if at all, does Heaps' law for LLM emulated corpora differ from the law as observed for associated human-authored texts?
    \item If valid, how, if at all, does Heaps' law change with increasing LLM model size?
\end{itemize}
In short, while we find clear evidence for Heaps' law in GPT-Neo emulated PubMed abstracts, vocabulary size grows at a faster rate than we observed in the human authored PubMed abstracts. Interestingly, the rate of vocabulary size growth slows with the size of the GPT-Neo model, suggesting that even larger models may reproduce Heaps' law as observed in human generated textual documents.

\begin{table*}[!ht]
\centering
\begin{tabular}{lccccccc}
\textbf{Corpus} & {$\hat\beta$} & {$\hat \alpha$} & {$r$} & {$V(N_d)$} & {$N_d$} & {$\bar k$} & {$w_1$}\\
\midrule
PubMed Abstracts & $0.6381 \pm 0.0000$ & $7.7972 \pm 0.0038$ & 0.9997 & 71,600,633 & 810,829   & 173 & 420,951 \\
GPT-Neo 125M     & $0.7924 \pm 0.0001$ & $1.1672 \pm 0.0013$ & 0.9989 & 64,109,196 & 1,834,958 & 154 & 1,482,751 \\
GPT-Neo 1.3B     & $0.7320 \pm 0.0001$ & $2.3558 \pm 0.0030$ & 0.9984 & 66,565,724 & 1,292,574 & 160 & 968,553 \\
GPT-Neo 2.7B     & $0.7232 \pm 0.0001$ & $2.6461 \pm 0.0031$ & 0.9986 & 64,618,857 & 1,213,606 & 156 & 904,344 \\
\midrule
\end{tabular}
\caption{Heaps' law estimated slope, $\hat \beta$, estimated intercept, $\hat \alpha$, and Pearson's correlation coefficient, $r$, for the corpora examined in this study accompanied by the corpus statistics $d$ (number of documents), $V(N_d)$ (vocabulary size), $N_d$ (collection size), $\bar k$ (average document length), and $w_1$ (number of singleton terms). The uncertainties in the $\beta$ and $\alpha$ estimates are 90\% confidence~bounds.}
\label{tbl:main-result}
\end{table*}

\subsection{Experimental Setup}
\noindent\textbf{Data Acquisition}: PubMed is an online archive for biomedical and life science article bibliographic information run by the National Library of Medicine~\cite{Wheeler2007}. Our experiments rely on data from the PubMed Abstracts corpus, a collection of abstracts from some $30$~million PubMed housed article bibliographic entries~\cite{Gao2020,Biderman2022}. PubMed Abstracts constitutes one of the $22$ constituent sub-corpora making up the $825$~GiB English text corpus known as the Pile~\cite{Gao2020,Biderman2022}, the corpus on which GPT-Neo models are trained. PubMed Abstracts is an ideal corpus on which to base our investigations on account that its constituent documents are short (i.e., average document length $\bar k = 173$) natural language texts. Computational cost and breakdowns in text coherence make emulating long texts comparatively problematic.

\noindent\textbf{Data Preprocessing}: We extracted the first $500,000$ PubMed Abstracts documents for our study. We settled on this data size to ensure that a broad range of biomedical topics are captured while keeping our computational workload to a manageable level. To achieve consistency in our dataset, we subjected the documents to a series of preprocessing steps. We implemented the procedure in Python~3.11.2. For each abstract, we first reduce all characters to their canonical Unicode representations, then converted all terms to lowercase, then removed all punctuation marks, then finally tokenized the resulting texts. To ensure quality and significance in our analysis, we removed all documents containing five words or fewer. This step was crucial to filter out any overly short or potentially irrelevant abstracts.

\noindent\textbf{Corpus Emulation}: We emulated the first 500,000 PubMed Abstract documents under each of GPT-Neo~125M, GPT-Neo~1.3B, and GPT-Neo~2.7B. For the generation of our emulated corpora, we adopted a straightforward prompting strategy. Specifically, we used the first five words of each PubMed abstract as a prompt. Rather than asking the GPT-Neo model to complete the text up to a maximum length of the original abstract, we employed a bucketing strategy. In this strategy, abstracts are grouped into buckets based on their lengths, ensuring efficient and balanced processing. Each bucket then provides a maximum allowable length for text completion. This method optimizes computational efficiency by preventing the model from over-generating on shorter prompts while also ensuring adequate generation length for longer prompts. It ensures that each emulated abstract starts with content relevant to the original document, providing a robust foundation for subsequent comparison of generated content.

\subsection{Experimental Results}
We assessed Heaps' law on the PubMed Abstracts corpus, as well as on each GPT-Neo generated corpus. Model parameters were estimated using ordinary least squares regression on the log-log transformed relation $\log_{10}(V(N_i)) = \beta\log_{10}(N_i) + \log(\alpha)$.

Figure~\ref{fig:heaps-law}A visualizes $N_i$ versus $V(N_i)$ on a log-log scale with a different colored curve for each corpus. That the curves trace out positively sloped straight lines is evident from visual inspection, indicating the corpora obey Heaps' law. Table~\ref{tbl:main-result} provides the estimated Heaps' law parameters for each analyzed corpus. For the PubMed Abstracts corpus, we find a linear fit, corresponding to the polynomial relation $V(N_i) = \alpha N_i^\beta$, with $\hat\beta = 0.6381 \pm 0.0000$ and $\hat\alpha = 7.7972 \pm 0.0038$. The correlation coefficient $r=0.9997$ suggests Heaps' law is a very good fit to the data. According to Baeza-Yates and Navarro~\cite{BaezaYates2004}, within English text corpora, the $\beta$ is typically between 0.4 and 0.6, and 10 and 100 for $\alpha$. The estimated exponent value for the PubMed Abstracts corpus falls beyond the upper endpoint of this reported typical interval of values, but not to an alarming extent. The slightly higher than typical rate of vocabulary growth might be attributable to the abstracts abounding in technical terminology and other jargon. The estimated $\alpha$ likewise lands a little outside the reported range. The GPT-Neo generated corpora are similarly observed to be in good agreement with Heaps' law. However, the emulated corpora manifest a comparatively high rate of vocabulary growth, all surpassing the mark of $0.70$ for $\hat\beta$.

Figure~\ref{fig:heaps-law}B depicts the same data on a natural scale to facilitate visual comparisons of vocabulary growth rates. Notably, an inverse relationship is observed between GPT-Neo model size and generated vocabulary size. The rate of vocabulary growth decreases with increasing model size. The GPT-Neo~125M model notably produced many singleton terms, a significant portion of which were not recognizable words. Similar trends were observed for the GPT-Neo~1.3B and~2.7B models, with both generating more singleton terms than are observed in the PubMed Abstracts corpus. We are left to conclude that it is high levels of singleton term generation, as compared with the human authored PubMed Abstracts documents, which is driving the high rates of vocabulary accrual in GPT-Neo emulated corpora.

In our exploration of Heaps' law in human-authored journal article abstracts compared with associated GPT-Neo-model-emulated abstracts, we found that
\begin{itemize}
    \item Corpora of GPT-Neo emulated abstracts follow Heaps' law.
    \item The emulated corpora exhibit significantly higher rates of vocabulary growth than are observed in typical human-authored corpora. This is explained by the GPT-Neo models penchant for producing many more singleton terms than we observed in the human data.
    \item The more complex the GPT-Neo model (in terms of the number of parameters), the more closely it mirrors the rates of vocabulary accrual typical of human generated text.
\end{itemize}
In summary, we present evidence that LLM generated text is in keeping with Heaps' law. However, the fewer the parameters on which the LLM is built, the more the rate of vocabulary growth outpaces that of typical human authored text.

\section{Conclusion}
In this short paper we present the first study of Heaps' law in LLM generated text. Our experiments with the GPT-Neo suite of models provides evidence that LLM text, like human produced text, follows Heaps' law. However, we found atypically high rates of vocabulary accrual in LLM generated text, which is attributable to the presence of many singleton terms. This effect is diminished by increasing the number of GPT-Neo model parameters.

We emphasize the preliminary status of this work. In the short-term, we envision strengthening the experimental findings presented here in a number of ways. First, we used a computationally conveniently sized sub-corpus of PubMed abstracts corpus. We are determined to analyze the full PubMed Abstracts corpus in subsequent experiments. We further aim to verify that Heaps' law estimated parameters are invariant under random shufflings of the documents. We will also explore more systematic approaches to prompting engineering when emulating PubMed abstracts~\cite{Ziems2023}. A crucial step will be the inclusion of GPT-NeoX~20B model~\cite{Black2022}. It is conceivable, given the 20 billion parameters, that GPT-NeoX~20B emulated PubMed abstracts exhibit Heaps' law exponents typical of human-authored text, but this is unknown. In the longer-term, we aim to extend our investigations of Heaps' law to other LLMs, including FLAN-T5~\cite{Chung2022} and LaMDA~\cite{Thoppilan2022}. We also aspire to extend the scope of this investigation to cover text from diverse domains and languages. Such expansion is crucial to ascertain the generalizability of Heaps' law in LLM emulated documents.

Lastly, the term \textit{memorization} describes the proclivity of LLMs to generate complete fragments of text exactly as they appear in their training data~\cite{Biderman2023}. The relation we observe between Heaps' law and model size invites the thought that the memorization abilities of LLMs might increase with model size. Further investigation is merited to elucidate connections between Heaps' law and the memorization phenomenon.

It is our hope that this study spurs further investigations into the corpora profiling of LLM generated~text.

\section*{Author Contribution Statement}
Uyen Lai: Writing – original draft and review \& editing, Methodology, Numerical experiments, Results interpretation, Computer code, Visualization. \\
Gurjit S. Randhawa: Writing – review \& editing, Results interpretation, Supervision, Funding acquisition. \\
Paul Sheridan: Writing - original draft and review \& editing, Conceptualization, Methodology, Results interpretation, Supervision, Funding acquisition.

\section*{Code Availability}
Project code available at the GitHub repository \url{https://github.com/paul-sheridan/paper-heaps-law-llm}, Release v1.

\begin{acks}
The work was funded by the Natural Sciences and Engineering Research Council of Canada Discovery Grant RGPIN-2022-03547 to Gurjit Randhawa, and University of Prince Edward Island Internal Research Grant 6011874 to Paul Sheridan. Advanced computing resources were provided by the Digital Research Alliance of Canada, the organization responsible for digital research infrastructure in Canada, and ACENET, the regional partner in Atlantic Canada.
\end{acks}

\bibliographystyle{ACM-Reference-Format}
\bibliography{bibliography}

\end{document}